%% file: iclr2021_conference.tex
\documentclass{article} % For LaTeX2e
\usepackage{iclr2021_conference,times}

\input{math_commands.tex}

\usepackage{hyperref}
\usepackage{url}
\usepackage{algorithm,algpseudocode}
\usepackage{amsmath}
\usepackage{times}  % DO NOT CHANGE THIS
\usepackage{helvet} % DO NOT CHANGE THIS
\usepackage{courier}  % DO NOT CHANGE THIS
\usepackage{graphicx} % DO NOT CHANGE THIS
\usepackage{booktabs}
\usepackage{multirow}
\usepackage{titlesec}
\titlespacing\section{0pt}{3pt plus 4pt minus 2pt}{0pt plus 2pt minus 2pt}

\title{Pairwise Relations Discriminator for \\
 Unsupervised Raven's Progressive Matrices}

\author{Nicholas Quek Wei Kiat, Duo Wang, Mateja Jamnik \\
Department of Computer Science \& Technology\\
University of Cambridge, UK\\

\texttt{nicholasquekweikiat@gmail.com,\{duo.wang,mateja.jamnik\}@cl.cam.ac.uk} \\
}

\iclrfinalcopy % Uncomment for camera-ready version, but NOT for submission.
\begin{document}
\maketitle

\begin{abstract}
The ability to hypothesise, develop abstract concepts based on concrete observations and apply these hypotheses to justify future actions has been paramount in human development. An existing line of research in outfitting intelligent machines with abstract reasoning capabilities revolves around the Raven's Progressive Matrices (RPM). There have been many breakthroughs in supervised approaches to solving RPM in recent years. However, this process requires external assistance, and thus it cannot be claimed that machines have achieved reasoning ability comparable to humans. Namely, humans can solve RPM problems without supervision or prior experience once the RPM rule that relations can only exist row/column-wise is properly introduced. In this paper, we introduce a pairwise relations discriminator (PRD), a technique to develop unsupervised models with sufficient reasoning abilities to tackle an RPM problem. PRD reframes the RPM problem into a relation comparison task, which we can solve without requiring the labelling of the RPM problem. We can identify the optimal candidate by adapting the application of PRD on the RPM problem. Our approach, the PRD, establishes a new state-of-the-art unsupervised learning benchmark with an accuracy of 55.9\% on the I-RAVEN, presenting a significant improvement and a step forward in equipping machines with abstract reasoning.
\end{abstract}

\section{Introduction}
Artificial general intelligence (AGI), that is, machines with the capacity to comprehend and execute any intellectual task which a human is capable of, is one of the goals of artificial intelligence (AI) research. However, the current state of AI is far from achieving AGI. One of the distinct characteristics of human intelligence is abstract reasoning: the ability to derive rules and concepts from concrete observations and apply logical reasoning to new observations in order to justify future actions (\cite{abstractthinking}). However, developing strong abstract reasoning capabilities alone in our machines is insufficient. These machines also need to be able to generalise their existing knowledge in order to develop new skills to solve new problems in new environments. Visual Questioning Answering (VQA) (\cite{vqa}) and Raven's Progressive Matrices (RPM) (\cite{rpm}) are existing lines of research that aim to equip machines with abstract reasoning capabilities. VQA evaluates capabilities lying on the periphery of the cognitive ability test circle such as spatial and semantic understanding (\cite{iqtest}). RPM tests one's joint spatial-temporal reasoning capabilities which form the core of human intelligence (\cite{iqtest}), making RPM a significantly more challenging task. A detailed description of RPM task can be found in Appendix~\ref{apx:rpm}.

Many supervised models have been proposed to solve RPM in recent years. SCL (\cite{scl}), the state-of-the-art model, demonstrated superhuman performance, achieving an accuracy of 95.0\% on the I-RAVEN dataset (\cite{iraven}). Despite these breakthroughs, machines still require human supervision to achieve this degree of reasoning ability, and are thus, still far from achieving reasoning ability comparable to humans. In particular, when the RPM rule that relations can only exist row/column-wise is properly introduced, humans can solve RPM problems without supervision or prior exposure. The only unsupervised approach to RPM currently is MCPT (\cite{mcpt}). It introduces the idea of transforming an unsupervised learning problem into a supervised one using a pseudo target. The model establishes the state-of-the-art for unsupervised approaches to RPM, with an accuracy of 28.5\% on the RAVEN dataset (\cite{drt}). 

We propose a novel approach, namely, the pairwise relations discriminator (PRD), to tackle this question. We reframe the RPM problem into a relation discrimination task. The function of the PRD is to determine whether two rows of three cells obey a common rule. To train this discriminator, we introduce the alternate relation generator (ARG). The ARG generates training samples, together with appropriate targets, using only information from the problem sample. Each training sample is a pair of rows labelled 1 if they originate from the same RPM problem and 0 otherwise. By reframing the problem, we obtain a labelled dataset which allows us to train the PRD in a fashion similar to supervised methods. To solve the original RPM problem, we offer some modifications to the inference process. We independently insert each candidate into the empty cell; then, using the PRD we score every pair consisting of a resultant row and each of the first two rows of the problem matrix; finally we select the highest-scoring candidate. 

To verify the effectiveness of the approach, we evaluated our models on the I-RAVEN test dataset. The PRD approach we proposed achieved a mean accuracy of 55.9\%, surpassing the previous best of 28.5\% by a significant margin. PRD also performs better on I-RAVEN than RAVEN despite the presence of a short-cut solution in RAVEN. This demonstrates that PRD does not exploit the statistical bias present in RAVEN.

% No longer relevant; the current SOTA in supervised learning behave the same way 
%During the evaluation, we observed that the models produced with unsupervised approaches displayed strengths and weaknesses in the same configurations as humans, seemingly mirroring human intelligence. 

% No longer relevant; I decided to train using train + val and evaluate on test
%We also obtained valuable insights into training data requirements. The inductive bias within the training dataset is not reflected in the model performance during the inference process. This allows us to use the test dataset to augment the training dataset.

\section{Method}
This section introduces our proposed approach to the RPM problem, the training process and how the model is adapted for inference. The proposed model is composed of two components, the Relations Extraction module and the Pairwise Relations Discriminator (PRD). The Relations Extraction module captures the relationship shared between the three cells from the same row while the PRD compares two relations to determine how similar they are.  

\textbf{Relation Extraction}: To extract the relation from a row of cells, we need a visual perception module to recognise the elements within a cell and a reasoning module to decipher the relationship between these elements across the row. For the visual processing component, we opted to use a convolutional neural network. In particular, ResNet-18 (\cite{resnet}) was selected because among the computer vision models explored in~\cite{drt}, ResNet had the best performance on the supervised RPM problem.  For our purposes, we combine the three single-channel images (a row of three greyscale cell images) into a single image (this is similar to MCPT, which proved successful). This change enables the model to perceive a row as an entity, not as three separate cells. It allows us to avoid directly addressing the short-term memory problem (where the elements of a prior cell are referenced when analysing another cell), and instead rely on the model to figure out the relationship between the channels.

\textbf{Pairwise Relations Discriminator}: Given two relations $r_i$ and $r_j$, the function of the PRD is to calculate the degree of similarity and return a similarity score, $s(r_i, r_j)$. The similarity score can be formulated as a function of a distance measure between relations $s(r_i, r_j) = f(d(r_i, r_j))$ where $d$ can be any distance measure. In this work we use  L1-distance measure for $d(r_i,r_j)$, as we found it performs best: $d(r_i, r_j) = |r_i - r_j|$. The distance feature is then fed into a dropout layer (\cite{dropout}), intended to reduce over-fitting. Next, an MLP with a single hidden layer of 128 dimensions is employed to produce a 1-dimensional output. The similarity score is obtained after normalisation with a sigmoid function, $\sigma$. As a result of normalisation, the similarity score is in the range of [0, 1] where 0 indicates no commonalities between the relations and 1 indicates that the two relations are identical:
\begin{equation} \label{eq:sigmamlp}
    s(r_i, r_j) = \sigma(MLP(d(r_i, r_j)))
\end{equation}

\textbf{Training}: Training targets are required to train the proposed model. We introduce the alternate relations generator (ARG) whose function is to generate `real' and `fake' data for training. The ARG takes an RPM problem and generates a pair of real and fake samples. Both real and fake samples each consist of two rows. The pair of rows in a real sample share a common relation, and hence have a target of 1. On the other hand, the two rows in the fake data have no common relation and are given a target of 0. The real and fake samples are related to positive and negative pairing in Noise Contrastive Learning (\cite{oord2018representation}). However, ARG uses a different pair sampling strategy, and has different training objective. To generate real data, the first two rows of the RPM are selected. These two rows are guaranteed to share the same relation since they belong to the same RPM problem. To bolster our model's ability to generalise, we shuffle the order of the rows. This subtle modification is important as it makes the model permutation-invariant, since an ideal RPM solver would not change its solution based on the row ordering of the problem.

\begin{algorithm}[t]
  \caption{Fake Data Generation}
  \label{fakealgorithm}
  \begin{algorithmic}
    \State \textbf{Input:} A RPM problem consisting of 8 question and candidate cells
    \State \textbf{Output:} A pair of rows, each containing 3 cells, with no common relation
    \State  
    \State i = 0 or 3 at random
    \State row\_1 = question\_cells[i : i+3]
    
    \If{random(0, 1) $\leq$ 0.5} \Comment{\textbf{Cat-A row}}
    \State problem2 = select a random RPM problem
    \State j = 0 or 3 at random
    \State row\_2 = problem2.question\_cells[j : j+3]
    \Else{} \Comment{\textbf{Cat-B row}}
    \State j = 3 - i
    \If{random(0, 1) $\leq$ 0.5} \Comment{$row_C$}
    \State row\_2 = question\_cells[6 : 8]
    \State row\_$\gamma$ = question\_cells[j : j+3]
    \State cell = select a random cell from row\_$\gamma$ 
    \Else{} \Comment{$row_\gamma$}
    \State row\_2 = question\_cells[j : j+2]
    \State cell = select a random cell from candidates
    \EndIf
    \State row\_2.append(cell)
    \State shuffle(row\_2)
    \EndIf
    \State\Return row\_1, row\_2
    \end{algorithmic}
\end{algorithm} 

The generation strategy for fake data is more involved and is illustrated in Algorithm \ref{fakealgorithm}. We label the three rows in a given RPM problem as $row_A$, $row_B$ and $row_C$. We first select either $row_A$ or $row_B$ and relabel it as $row_1$ with the unselected row relabelled as $row_\gamma$. To have a target of 0, the second row in the fake data sample must not share the same relation as $row_1$. There are many ways of obtaining such a second row. We decided to use rows from two categories. The first category (Cat-A) are rows with a completely different relation. We randomly select a different RPM problem and a row in the first two rows of that problem is randomly chosen as $row_2$. Given the large rule space, the probability of picking a row with the same relation is negligible. The second category (Cat-B) are rows with similar visual elements, but they do not necessarily share or contain a rule. To generate a Cat-B row, we pick at random from $row_C$ and $row_\gamma$. If the selected row is $row_C$, we randomly fill the absent third cell with a cell from $row_\gamma$. If $row_\gamma$ is picked, we retain the first two cells and replace the last cell with a candidate from the answer set. In both cases, once we obtain all three cells for our row, we shuffle these cells to eliminate any relation, and thus obtain a Cat-B row. Using this method, a Cat-B row will have similar visual elements since the cells come from the same RPM problem, but the probability of the relation being the same is extremely slim. The ARG we employ generates fake data with Cat-A and Cat-B rows at a ratio of 1:1. While there is a small chance that the sampled fake data contains rows of the same relation, this type of labelling noise (\cite{oord2018representation,rolnick2017deep}) is shown to have minimal effect on neural network training.

\begin{figure*}[t]
\centering
\includegraphics[width=\linewidth]{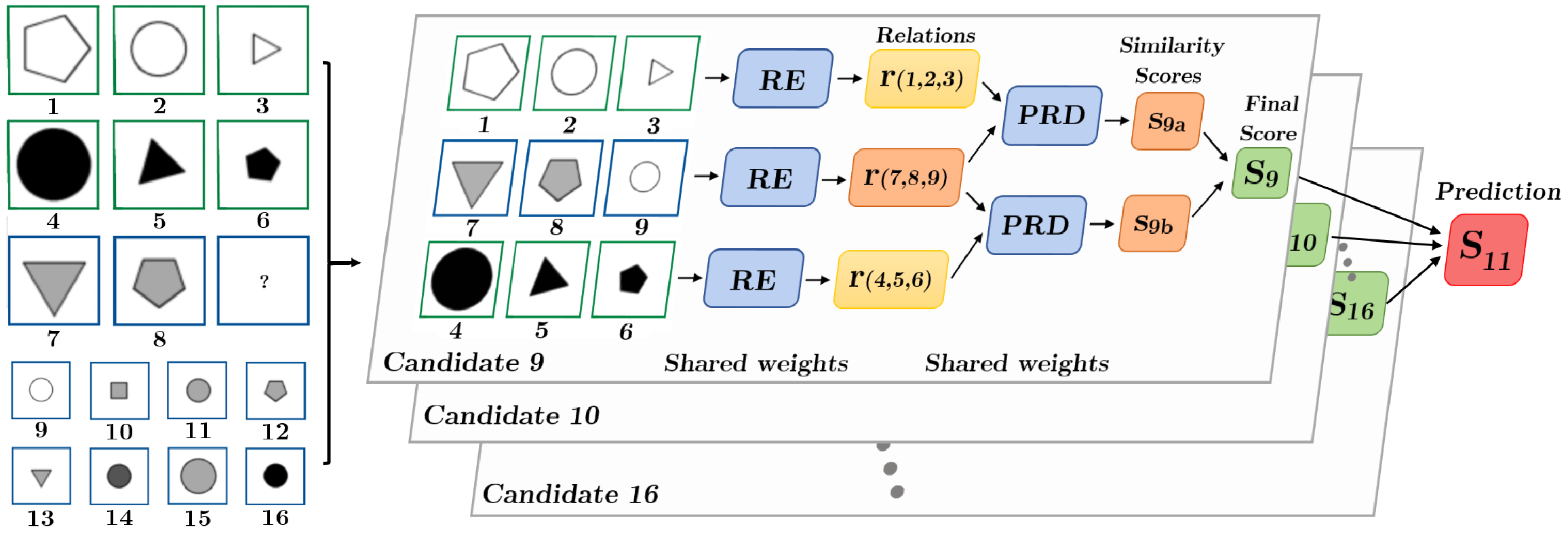}
\caption{The given RPM problem is deconstructed into eight row-triplets, one for each candidate. Each row is processed by the relation extractor (RE) to give a relation. The relations are pairwise processed by the PRD to produce a similarity score. The two similarity scores are combined to give a final score. The highest scoring candidate is the predicted candidate.}
\label{fig:inference}
\vspace{-0.4cm}
\end{figure*}

\textbf{Inference}: Our goal is to solve RPM problems. Therefore, we need to modify the input slightly, in order to utilise our model for inference on RPM problems. Figure \ref{fig:inference} 
visualises the inference process. We begin by deconstructing the RPM problem into eight row-triplets. Suppose the cells in our problem matrix are labelled 1-8 and the candidates in the answer set are labelled 9-16. $row_{i,j,k}$ is a row containing cells $i$, $j$ and $k$ in that order. A row-triplet contains the first two rows of the RPM problem matrix, $row_{1,2,3}$ and $row_{4,5,6}$. The final row of a row-triplet is constructed by inserting a candidate, $i \in [9, 16]$, into the final row of the problem matrix to create $row_{7,8,i}$. Since there are eight candidate answers, we have eight row-triplets. For each row-triplet, the relation extractor (RE) is applied to each row, $row_{i,j,k}$, to produce a relation, $r_{i,j,k} = RE(row_{i,j,k})$. By pairing each of the first two rows with the final row, $row_{7,8,i}$, we create two row-pairs. Each row-pair is passed through the PRD and so the similarity scores, $s_{ia}$ and $s_{ib}$ are produced:
\begin{equation}
s_{ia} = PRD(r_{1,2,3}, r_{7,8,i}), \quad s_{ib} = PRD(r_{4,5,6}, r_{7,8,i})
\end{equation}
The final score for that candidate cell, $S_i$, is the mean of the two similarity scores: $S_i = \tfrac{1}{2} (s_{ia} + s_{ib})$. Each row-triplet undergoes the same process and has a final score associated with it. The candidate corresponding to the triplet with the highest score is the model prediction. 

\section{Evaluation}
\begin{table*}[h!]
\centering
\normalsize
\begin{tabular}{@{}llcccccccc@{}}
\toprule
Method                        & \multicolumn{1}{c}{}           & Avg            & Center         & 2x2Grid        & 3x3Grid        & L-R            & U-D            & O-IC           & O-IG           \\ \midrule
\multirow{3}{*}{Supervised}   
& \multicolumn{1}{l}{CoPINet} & 46.3 & 54.4 & 33.4 & 30.1 & 56.8 & 55.6 & 54.3 & 39.0 \\
& HriNet & 63.9 & 80.1 & 53.3 & 46.0 & 72.8 & 74.5 & 71.0 & 49.6 \\ 
& SCL & \textbf{95.0} & \textbf{99.0} & \textbf{96.2} & \textbf{89.5} & \textbf{97.9} & \textbf{97.1} & \textbf{97.6} & \textbf{87.7} \\ \midrule
\multirow{3}{*}{Unsupervised} 
& Random & 12.5 & 12.5 & 12.5 & 12.5 & 12.5 & 12.5 & 12.5 & 12.5 \\
& MCPT\footnotemark[1] & 28.5 & 35.9 & 26.0 & 27.2 & 29.3 & 27.4 & 33.1 & 20.7 \\
& PRD (ours) & \textbf{55.9} & \textbf{73.1} & \textbf{39.9} & \textbf{35.3} & \textbf{67.3} & \textbf{67.3} & \textbf{68.1} & \textbf{40.6}  \\ \midrule
Human\footnotemark[1] & & 84.4 & 95.5 & 81.8 & 79.6 & 86.4 & 81.8 & 86.4 & 81.8 \\ \bottomrule
\end{tabular}
\caption{Test accuracy of each model on the I-RAVEN dataset.}
\label{tab:baselines}
\end{table*}
\footnotetext[1]{Evaluated on the original RAVEN dataset.}

The PRD model is trained on 56,000 samples from the I-RAVEN training and validation dataset and its performance is measured on the 14,000 samples in the test set. For comparison, we report several available results from both supervised and unsupervised approaches to solving the RPM task. Table~\ref{tab:baselines} presents the test accuracy of various models on the I-RAVEN dataset for the different  configurations (for details, see Table~\ref{fig:configurations} in Appendix~\ref{apx:rpm}). The first part of Table~\ref{tab:baselines} shows the test accuracy of the current top-performing supervised models: CoPINet (\cite{copinet}), HriNet (\cite{iraven}) and SCL (\cite{scl}). The second part reports on the results of the only unsupervised model, MCPT. We include a baseline of random guessing. Since there are 8 candidate answers for a given RPM problem, the average accuracy of a random guessing strategy is 12.5\%. Finally, we also include the performance of humans (\cite{drt}) for comparison. 

Despite being unsupervised, PRD outperforms the supervised CoPINet method on all configurations by an average of 9.6\%. Since MCPT was evaluated on the original RAVEN dataset, we cannot compare it with PRD directly. For completeness, we present the performance of PRD on the original RAVEN dataset and compare them in Appendix~\ref{apx:ravenperf}. Interestingly, PRD performs better on I-RAVEN than RAVEN despite the short-cut solution being eliminated from I-RAVEN. This demonstrates that PRD does not exploit the statistical bias present in RAVEN.

\pagebreak
\bibliography{iclr2021_conference}
\bibliographystyle{iclr2021_conference}
\pagebreak
\appendix
\section{Background}

\subsection{Raven Progressive Matrices}\label{apx:rpm}

\begin{figure}[h]
\centering
\includegraphics[width=0.5\linewidth]{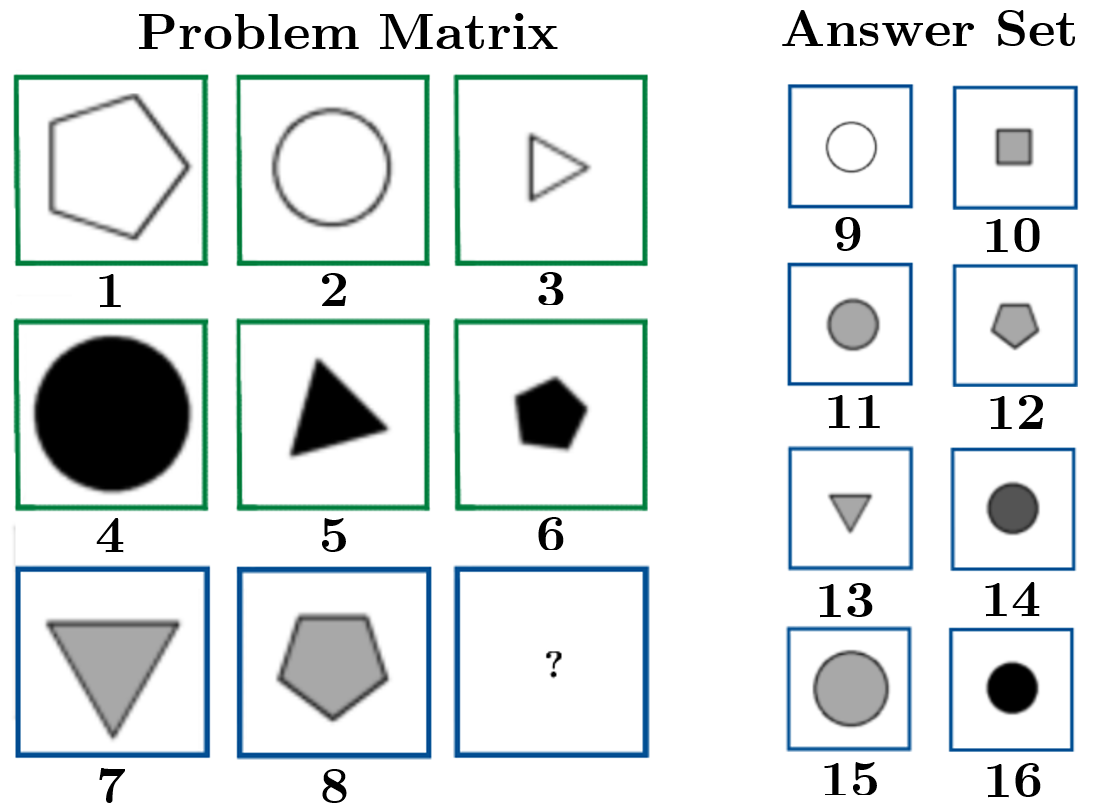}
\caption{A RPM problem from the RAVEN dataset \cite{drt}. The hidden logical rules, pertaining to \textit{shape, size} and \textit{colour}, are applied row-wise in this problem. The only candidate which satisfies all three rules is 11.}
\label{fig:problem}
\end{figure}

RPM is a test of abstract reasoning and fluid intelligence (\cite{fluidintel}). It presents a non-verbal question, consisting of a 3 x 3 matrix with one empty cell and an answer set of 8 alternative cells to fill the empty cell (see Figure~\ref{fig:problem}). Each cell contains visually simple elements, which when viewed as an entire matrix, obey a specific rule. This non-reliance on language makes it applicable as a mean of assessment across populations from different languages, with varying reading and writing skills, as well as of different cultural backgrounds. RPM has been shown to be strongly diagnostic of abstract and structural reasoning ability, capable of discriminating even among highly educated populations (\cite{diagnostic}). These properties have, over the years, propelled RPM as a leading test for Intelligence Quotient (IQ) of humans (\cite{iqtest}). To solve a RPM, we need to derive the rule with which the matrix was constructed, which typically involves sophisticated logic, including recursions. The rule may be composed of different sub-rules at various levels in its structure, making the reasoning process extremely difficult. Derivation of the rule requires joint spatial-temporal reasoning across both the problem matrix and the answer set (\cite{iqtest}), which involves visual processing, short term memory, sequential and inductive reasoning. To acquire these capabilities in machines, both perception and reasoning subsystems are necessary.

The first large-scale RPM dataset, Procedurally Generated Matrices (PGM) (\cite{pgm}), was introduced in 2018, sparking machine learning interest on the topic. Subsequently, the Relational and Analogical Visual rEasoNing (RAVEN) (\cite{drt}) dataset was developed to include structure and hierarchy which were absent in PGM. 

\begin{figure*}[t!]
\centering
\includegraphics[width=\linewidth]{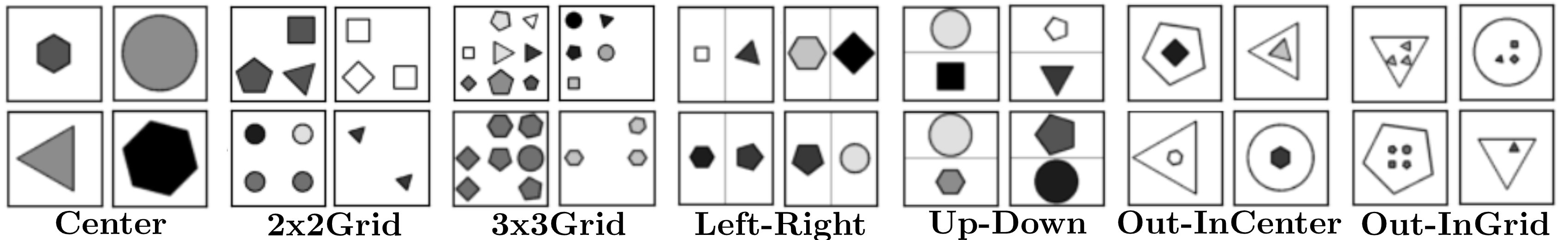}
\caption{The 7 distinct figure configurations present in RAVEN. In \textit{Center}, a single shape occupies the entire cell. A grid layout is present in \textit{2x2Grid} and \textit{3x3Grid}, where each section may house a single shape. In \textit{Left-Right} and \textit{Up-Down}, the cell is split into two halves vertically and horizontally, where each half contains a single shape. For \textit{Out-InCenter} and \textit{Out-InGrid}, the cell is split into an outer and inner component. The outer component contain a shape while the inner component follows the layout of \textit{Center} and \textit{2x2Grid} respectively.}
\label{fig:configurations}
\end{figure*}

As a tool for measuring reasoning capabilities, the visual recognition task was kept simple. Each cell contains a small set of simple clearly defined grey-scale elements without any occlusions. The rules, however, were intricately crafted to present a cognitive challenge that best measures reasoning ability. RAVEN employs a system of attribute-rule pairs, where each attribute (\textit{Position}, \textit{Type}, \textit{Size} and \textit{Colour}) is matched with a rule. There are four distinct categories of rules, which are (\textit{Constant}, \textit{Progression}, \textit{Arithmetic}, \textit{Distribute Three}). To increase the difficulty of the problem, two additional attributes are implemented as noise attributes, \textit{Uniformity} and \textit{Orientation}, to misdirect the solver. The system of attribute-rule pairs is enforced row-wise. For details of attributes and rules in RAVEN dataset, please refer to~\cite{drt}. 
RAVEN establishes 7 distinct configurations which are shown in Figure~\ref{fig:configurations}. The average number of non-constant rules in a problem's rule system is 6.29, providing a challenge even for a competent solver. 

RPM problems are constructed by first sampling a rule system. Visual elements with attributes which conform to the rule system are then selected. Through this process, the RAVEN dataset was developed. The dataset contains 1,120,000 images organised into 70,000 problems, distributed equally across the 7 distinct configurations. The authors (\cite{drt}) split the dataset into 3 components. 20\% of the data was set aside as a held-out test set. The remaining data is further split into a training set and a validation set at a ratio of 3:1. The authors also collected a human-level performance baseline (\cite{drt}). Human participants consisting of college students (from UCLA) were evaluated on a subset of representative samples from the dataset. Participants were first familiarised with RPM problem with only one non-constant rule in a fixed configuration. After familiarisation, participants were assigned problems with complex rule combinations, and their answers were recorded. With a human-level performance baseline, the RAVEN dataset can serve as a benchmark to measure the reasoning ability of machines.

\subsection{I-RAVEN}\label{apx:iraven}
\cite{iraven} observed that there are defects in the design of candidate set generation in the RAVEN dataset. They discovered that the correct answer can be found by simply scanning the answer set without any information of the context images. This short-cut solution goes against the essence of abstract reasoning, undermining the RAVEN dataset's ability to evaluate abstract reasoning ability. Hence, they introduced a revised dataset, Impartial-RAVEN (I-RAVEN), which eliminates the short-cut solution by generating the candidate set with a different algorithm. Therefore, in this paper, we compared our results on the I-RAVEN dataset. Results on the original RAVEN are available in Appendix \ref{apx:ravenperf} for completeness.

\section{Hyperparameters}
All models were implemented in PyTorch, trained and evaluated on a single GPU of NVIDIA TITAN X Pascal. The ResNet-18 network is pre-trained using the ImageNet dataset. To maximise the effects of the pre-training, our input is preprocessed to match the ImageNet dataset. All images are resized to a resolution of 224 x 224 pixels. The pixel values are rescaled to the range of [0, 1], standardised with the means (0.485, 0.456, 0.406) and standard deviations (0.229, 0.224, 0.225) of the RGB channels of the ImageNet dataset. All batch normalisation layers within the network were frozen. For the PRD module, the dropout layer was set to 0.5. Data was organised into mini-batches of 32. Each mini-batch consisted of only real or only fake pairs. The binary cross entropy (BCE) loss function was used to compute gradients for backpropagation. The mean gradients computed from a real and a fake mini-batch were used to update the model parameters with the use of the Adam optimiser (\cite{adam}) with a fixed learning rate of 0.0002. 

Ideally, we would use a metric which can be calculated without labelled data as an indicator to stop training. However, we were unable to find a metric which strongly correlates with model performance, despite experimenting with multiple metrics. Hence, we decided instead to use the training loss as the indicator. Once the loss begins to plateau, we randomly select five checkpoints within that plateau. The average performance of these checkpoints on the test set are then reported.

\section{Performance on RAVEN} \label{apx:ravenperf}
\begin{table*}[t]
\centering
\normalsize
\begin{tabular}{@{}lcccccccc@{}}
\toprule
Model                      & Avg            & Center         & 2x2Grid        & 3x3Grid        & L-R            & U-D            & O-IC           & O-IG           \\ \midrule
MCPT & 28.5 & 35.9 & 26.0 & \textbf{27.2} & 29.3 & 27.4 & 33.1 & 20.7 \\
PRD (ours) & \textbf{37.9} & \textbf{57.8} & \textbf{26.8} & 24.8 & \textbf{43.4} & \textbf{43.3} & \textbf{46.7} & \textbf{22.9} \\
\bottomrule
\end{tabular}
\caption{Evaluated on the RAVEN dataset.}
\label{tab:raveneval}
\end{table*}

Table \ref{tab:raveneval} reports the performance of the unsupervised models on the RAVEN test dataset. The testing accuracy of PRD is significantly better than MCPT, the only other unsupervised approach. PRD outperforms MCPT on every configuration except for \textit{3x3Grid}. In particular, for the \textit{Center} configuration, the margin in performance is 21.9\%. However, the margin is not as wide for the more challenging configurations (\textit{2x2Grid} and \textit{Out-InGrid}). 

Table \ref{tab:prdeval} collates the performance of PRD on the two datasets. Interestingly, PRD performs better on I-RAVEN than RAVEN despite the short-cut solution being eliminated from I-RAVEN. This demonstrates that PRD does not exploit the statistical bias present in RAVEN. %In RAVEN, all 7 distractors differ from the solution by exactly one attribute. In I-RAVEN, only 3 distractors differ by exactly one attribute; the other differ by a greater number of attributes (2, 2, 2, 3). If we do not consider the statistical bias in the answer panel, I-RAVEN is the easier dataset.

\begin{table*}[t]
\centering
\normalsize
\begin{tabular}{@{}lcccccccc@{}}
\toprule
Dataset                      & Avg            & Center         & 2x2Grid        & 3x3Grid        & L-R            & U-D            & O-IC           & O-IG           \\ \midrule
RAVEN & 37.9 & 57.8 & 26.8 & 24.8 & 43.4 & 43.3 & 46.7 & 22.9 \\
I-RAVEN & \textbf{55.9} & \textbf{73.1} & \textbf{39.9} & \textbf{35.3} & \textbf{67.3} & \textbf{67.3} & \textbf{68.1} & \textbf{40.6}  \\
\bottomrule
\end{tabular}
\caption{Performance of PRD on the different datasets.}
\label{tab:prdeval}
\end{table*}

\end{document}

%% file: math_commands.tex
\usepackage{amsmath,amsfonts,bm}

\def\eqref#1{equation~\ref{#1}}
\def\1{\bm{1}}

\DeclareMathAlphabet{\mathsfit}{\encodingdefault}{\sfdefault}{m}{sl}
\SetMathAlphabet{\mathsfit}{bold}{\encodingdefault}{\sfdefault}{bx}{n}